\documentclass{llncs}
\usepackage{graphicx}
\usepackage[english]{babel}
\usepackage{amsmath}
\usepackage{amsbsy}
\usepackage{amssymb}
\usepackage{amsfonts}
\usepackage{epsfig}

\begin{document}

\pagestyle{empty}

\mainmatter

\title{Evolving Genes to Balance a Pole}

\author{Miguel Nicolau\inst{1} \and Marc Schoenauer\inst{1} \and Wolfgang Banzhaf\inst{2}}
\institute{INRIA Saclay - \^Ile-de-France\\
LRI- Universit\'e Paris-Sud, Paris, France\\
\email{\{Miguel.Nicolau,Marc.Schoenauer\}@inria.fr}
\and
Memorial University\\
Newfoundland, Canada\\
\email{wolfgang@mun.ca}
}
\date{\today}  
\maketitle

\begin{abstract}
We discuss how to use a Genetic Regulatory Network as an
evolutionary representation to solve a typical GP reinforcement problem, 
the pole balancing. The network is a modified
version of an Artificial Regulatory Network proposed a few
years ago, and the task could be solved only by finding a
proper way of connecting inputs and outputs to the network.
We show that the representation is able to generalize well
over the problem domain, and discuss the performance of
different models of this kind.
\end{abstract}

\section{Introduction}

Knowledge of biological systems has come a long way since the inception of
the evolutionary computation field \cite{banzhaf06a}. Their remarkable flexibility 
and adaptivity seems to suggest that more biologically based
representations could be applied as representations for program evolution, i.e., Genetic Programming (GP).
The objective of this paper is exactly that - to apply a recent biological
model as a basis for some GP representations. 

We are interested here in a complexification
of the genotype-phenotype mapping, a process that seems to contribute
to a higher evolvability of genomes \cite{kirschner05}. A central piece 
of this mechanism is the regulation of genes by genes, what has become
known as Genetic Regulatory Networks (GRNs).

GRNs are biological inter\-action networks among the genes in a chromosome 
and the proteins they produce: each gene encodes specific types of protein, and 
some of those, termed \textit{Transcription Factors}, regulate (either enhance or 
inhibit) the expression of other genes, and hence the generation of the protein those 
genes encode. The study of such networks of interactions provides many
interdisciplinary research opportunities, and as a result, GRNs have become an
exciting and quickly developing field of research \cite{hasty01}.

The question of how to use a GRN approach for GP is a
challenge that is being recognized only slowly by GP researchers. While some
progress has been made \cite{lones04a,song08}, there is yet to be proposed a proper unification of
the counteracting tendencies of networks to produce dynamics and continuous
signals versus the boolean logic and operator-operand-based methodology of traditional
GP. 

In this contribution, we shall study whether and how
the Artificial Gene Regulatory Model proposed in \cite{banzhaf03a} can be used 
to achieve the function traditionally implemented by control algorithms, by applying
it to a classical benchmark problem of control engineering, pole balancing. 

Along the way, we hope to learn how to use this type of representation for
problems usually solved with less evolvable representations. Our goal is 
to arrive at a flexible and at the same time very general representation useful
in GP in general. While we are not there yet, we have made progress notably
by finding ways to couple input and output to artificial GRNs, a feature of
utmost importance in Genetic Programming. 

This paper is organised as follows. Section \ref{sec:model} describes the GRN
model used, along with an analysis of its behaviour and modifications done
in order to adapt it to the evolution of solutions for typical GP problems.
Section \ref{sec:pole} then describes the problem and the evolutionary
algorithm we shall use to solve it.  Section \ref{sec:results} describes some
of the experiments conducted, and finally Section \ref{sec:conclusions} draws
conclusions and discusses future work directions.




\section{Artificial Gene Regulatory Model}
\label{sec:model}

\subsection{Representation and dynamics}
\label{promoter}
The model used in this work \cite{banzhaf03a} is composed of a
genome, represented as a binary string, and mobile proteins, which interact with
the genome through their binary signatures: they do so at
\textit{regulatory sites}, located upstream from genes. The resulting
interaction regulates the expression of the associated gene.

Genes are identified within the genome by \textit{Promoter sites}. These
consist of an arbitrarily selected 32 bit bit pattern:
the sequence \texttt{XYZ01010101} identifies a gene, with \texttt{X},
\texttt{Y} and \texttt{Z} representing each an arbitrary sequence of 8 bits.

If a promoter site is found, the 160 bits ($5\times32$)
following it represent the gene sequence, which encodes a protein. This
protein (like all others in the model) is a 32 bit sequence, resulting
from a many-to-one mapping of the gene sequence: each bit results from
a majority rule for each of the five sets of 32 bits.

\begin{figure}[ht]
	\begin{center}
		\includegraphics[width = 0.8\textwidth]{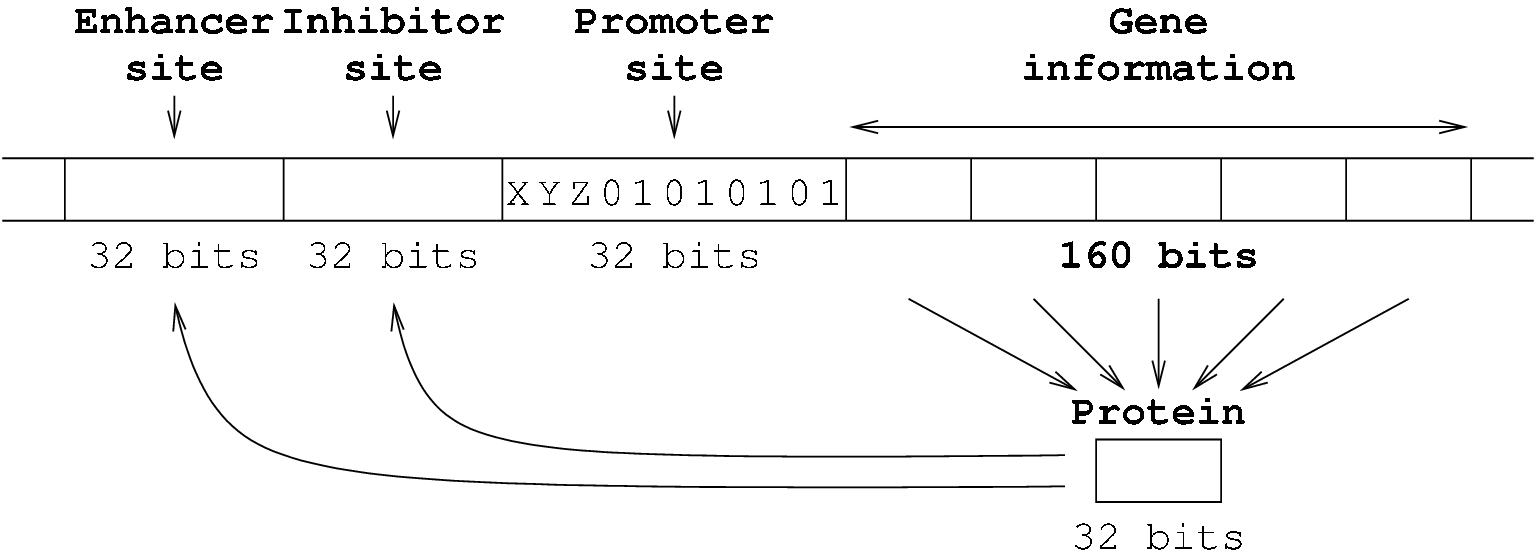}
	\end{center}
	\caption{Bit string encoding of a gene. If a promoter site is found,
	the gene information is used to create a protein, whose quantity is
	regulated by the attachment of proteins to the enhancer and inhibitor
	sites.}
	\label{fig:gene}
\end{figure}

Upstream from the promoter site exist two additional 32 bit segments,
representing the \textit{enhancer} and \textit{inhibitor} sites: these regulate
the protein production of the associated gene. The attachment of proteins to
these regulatory sites is what regulates this production. Fig.~\ref{fig:gene}
illustrates the encoding of a gene.

The binding of proteins to the regulatory sites is calculated
through the use of the \texttt{XOR} operation, which returns the degree of
match as the number of bits set to one (that is, the number of complementary
bits between both binary strings).

The enhancing and inhibiting signals regulating the production of protein $p_i$
are calculated by the following equation:
\begin{equation}
	e_i,h_i=\frac{1}{N}\sum^{N}_{j=1}c_j\exp(\beta(u_j-u_{max}))
	\label{eq:e_i}
\end{equation}
\noindent where $N$ is the total number of proteins, $c_j$ is the
concentration of protein $j$, $u_j$ is the number of complementary bits between
the (enhancing or inhibitory) regulating site and protein $j$, $u_{max}$ is the
maximum match observed in the current genome, and $\beta$ is a positive scaling
factor. Because of the exponential, only proteins whose match is close to $u_{max}$ will have an influence here.

The production of $p_i$ is calculated via the
following differential equation:
\begin{equation}
	\frac{dc_i}{dt}=\delta(e_i-h_i)c_i-\Phi(1.0)
	\label{eq:dc_i}
\end{equation}
\noindent where $\delta$ is a positive scaling factor (representing a time
unit), and $\Phi(1.0)$ is a term that proportionally scales protein production,
ensuring that $\sum_i{c_i} = 1.0$, which results in competition between binding
sites for proteins.

\subsection{Initialisation}
\label{sec:init}

Genomes can be initialised either randomly, or by using a duplication and
mutation technique \cite{banzhaf03a}: it consists in creating a random 32 bit
sequence, followed by a series of length duplications with a typical low
mutation rate associated. It has been shown \cite{wolfe97a,kellis04a} that
evolution through genome duplication and subsequent divergence (mostly
deletion) and specialisation occurs in nature.

In the following, genomes that have been initialized using a sequence of
Duplications and Mutations will be termed ``DM-genomes'' by contrast to the
``random genomes''.

\subsection{Input / Output}
\label{sec:io}

Most GP-like problems associate a given set of input values with a set of
responses (or outputs), and then measure the fitness of an individual as the
difference between the responses obtained and the known correct outputs.
However, the model as presented in \cite{banzhaf03a} is a closed world. 
This Section will now extend it with
I/O capabilities, so that it can be applied to typical GP problems.

\subsubsection{Model Input}

In order to introduce the notation of an input signal, the current model was
extended through the insertion of \textit{extra proteins}: regulatory
proteins not produced by genes, which are inserted into the model at a given
time.

Like the proteins which are produced by the genes in the model, these are also
$32$-bit binary strings, and like the other regulatory proteins, they cooperate
in the regulation of the expression of all genes, through the application of
Eq.~\ref{eq:e_i}. However, since they are not produced by specific genes, their
concentration is always the same across time (unless intentionally modified,
see below).

As these are regulatory proteins, their concentration is considered to take up
part of the regulatory process. This means that the differential equation used
(Eq.~\ref{eq:dc_i}) to calculate the expression level of TR-genes is changed as
follows:
\begin{equation}
	\frac{dc_i}{dt}=\delta(e_i-h_i)c_i-\Phi(1.0 - \sum_{j=N+1}^{N_{ep}}{c_j})
	\label{eq:newdc_i}
\end{equation}
\noindent where $N+1, \ldots, N_{ep}$ are the indices of the extra proteins in the model, and
$\Phi(1.0 - \sum_{j=N+1}^{N_{ep}}{c_j})$ is a term that proportionally scales
protein concentrations, such that the sum of all protein concentrations (gene
expression and extra proteins) adds up to $1.0$.

These extra proteins can be associated with problem inputs in two ways:
\begin{itemize}
	\item The binary signatures of the proteins represent the input values;
	\item The concentrations of the proteins represent the input values.
\end{itemize}

Each has its advantages and disadvantages. Setting binary signatures allows
evolution to exploit binary mutation to find useful matches between binary
signatures, but has a low resolution for continuous domains.  Setting
quantities is more adequate to represent continuous domains, but can be hard to
tune - a low extra protein concentration will hardly influence the regulatory
process, whereas a high concentration might crush the role of TF-genes.

\subsubsection{Model Output}

As mentioned before, each gene in the model encodes a transcription factor,
which is used in the regulatory process. In nature, however, these are only a
subset of the proteins expressed by genes. One could have proteins
with different roles in the model, and use some as outputs of the
model.

Keeping this idea in mind, the model has been adapted, so that different kinds
of promoters can be detected, to identify different types of gene. This allows
one to give specific roles to the proteins produced by each type of gene.

In this work, two types of genes were identified in the model: genes encoding
transcription factors (\textit{TF-genes}) and genes encoding a \textit{product
protein} (\textit{P-genes}). The first ones act just like in the original model \cite{banzhaf03a}: their
proteins regulate the production of all genes, regardless of their type. The
second ones are only regulated: their actual output signal is left for
interpretation to the objective function.
In order to identify different types of genes, the genome is scanned for
different promoter sites. Dropping the ambiguous sequence used in the original model (see Section \ref{promoter}), the following binary sequences were used: 
\texttt{XYZ00000000} to identify TF-genes, and \texttt{XYZ11111111} to identify
P-genes, as they have both the same probability of appearing (and no
overlapping of their signatures).

Note that a previous approach for extracting an output signal from this model exists
\cite{kuo04a}, where a random site of the genome is used as a regulation site, but
despite the results achieved, it does not offer the same degree of flexibility as the
technique now presented.

\subsubsection{Dynamic analysis}

Several possibilities exist, when choosing the dynamic equation to use when
calculating the concentration of P-proteins. In order to keep with the nature of the
model, equations based on the calculation of concentration of TF-proteins were
tested; the following equation was used:
\begin{equation}
	c_i^t = c_i^{t-1} + \delta(e_i - h_i)-\Phi(1.0)
\end{equation}
\noindent where $c_i^t$ is the concentration of the P-protein at time $t$, $c_i^{t-1}$ its
concentration at time $t-1$, $e_i$ and $h_i$ are calculated as before at time $t-1$,
and $\Phi(1.0)$ is a scaling factor, ensuring the sum of all 
P-proteins concentrations\footnote{Concentrations of TF-proteins and P-proteins are normalised independently.} is $1.0$.

This equation was chosen as it seems to give P-genes similar dynamics to
TF-genes, for both random genomes and DM-genomes 


\section{The Problem: Single-Pole Balancing}
\label{sec:pole}

The potential of using gene regulatory networks as a representation for an
Evolutionary Algorithm lies in their possibly rich, non-linear dynamics
\cite{kuo04a}. A famous dynamic control benchmark is the pole-balancing problem
\cite{barto83a,whitley93a}, also known as the inverted pendulum problem. It
consists in controlling, along a finite one dimensional track, a cart on which
a rigid pole is attached. The command is a bang-bang command: the user can
apply a constant force to either side of the cart. The objective is to keep the
pole balanced on top of the cart, while keeping the cart within the (limited)
boundaries of the track.

\noindent There are four input variables associated with this problem:\\
\indent $x \in [-2.4, 2.4]\ m$ is the position of the cart,
relative to the centre;\\
\indent $\theta \in [-12, 12]\ ^\circ$ is the angle of the pole with the
vertical;\\
\indent $\dot{x} \in [-1, 1]\ m/s$ is the velocity of the cart on the track;\\
\indent $\dot{\theta} \in [-1.5, 1.5]\ ^\circ/s$ is the angular velocity of the pole.\\

The physical simulation of the cart and pole movements is modelled by the
following equations of motion:
\begin{align*}
\ddot{\theta}(t) &= \frac{g \sin \theta(t) - \cos \theta(t) \left(\frac{F(t) + m l\dot{\theta}(t)^2\sin\theta(t)}{m_c + m}\right)}
		{l \left(\frac{4}{3} - \frac{m \cos^2\theta(t)}{m_c + m}\right)}\\
\ddot{x}(t) &= \frac{\frac{F(t) + m l\dot{\theta}(t)^2\sin\theta(t)}{m_c + m} - m l \ddot{\theta(t)} \cos{\theta(t)}}
		{m_c + m}
\end{align*}
\noindent where $g = 9.8\ m/s^2$ is the gravity, $l = 0.5\ m$ the half-pole length, $F(t) = \pm10\ N$ is the bang-bang command allowed, $m = 0.1\ \textrm{kg}$ and $m_c = 1.0\ \textrm{kg}$ are the masses of the pole and the cart respectively.\\

A time step of $0.02s$ is used throughout the simulations. A failure signal is
associated when either the cart reaches the track boundaries ($x=\pm2.4m$), or
the pole falls (i.e., $|\theta| > 12^\circ$).

The resulting controller accepts the four inputs, and outputs one of two
answers: push the cart left or right (with constant force $F(t) = \pm10N$).

\subsection{Encoding the Problem}

The four inputs were encoded using extra proteins, as explained in Section
\ref{sec:io}. These had the following signatures:\\
$x$: \texttt{\small 00000000000000000000000000000000}$\qquad\theta$: \texttt{\small 00000000000000001111111111111111}\\
$\dot{x}$: \texttt{\small 11111111111111110000000000000000}$\qquad\dot{\theta}$: \texttt{\small 11111111111111111111111111111111}\\

\noindent They were chosen such that their signatures are as
different as possible. Their concentration dictates their value: each of them
had the corresponding value of the input variable, scaled to the range
$[0.0,0.1]$. This means that the cumulated regulatory influence of these extra proteins
ranged from $0\%$ up to $40\%$.

The GRN was allowed to stabilize first, and then tested against a random
cart state, as seen in the literature. This is thus a very noisy fitness
function, as several combinations of the four input variables result in
unsolvable states (i.e.~the pole cannot be balanced). Success is dictated by a
successful series of $120000$ time steps without the cart exiting the $\pm2.4m$
track, or the pole falling beyond the $\pm12^\circ$ range. The (minimising)
fitness is thus:
$$
F(x) = \frac{120000}{\textrm{sucessful time steps}}
$$

The output action extracted from the genome is the concentration of a single
P-protein: a concentration above $0.5$ pushes the cart right, and vice-versa.
In the current work, all P-genes that are present in the genome are tested, 
and the most successful one is used.

As relevant concentration must be close to $0.5$, small genomes were used (the
higher the number of P-genes, the lower the probability of having a $0.5$ P-protein
concentration). The genomes were hence initialised with only $7$ DM events, with
$2\%$ mutation rate, generally leading to very small genomes.

As an alternative to this approach, another technique was used, which consists
in extracting the derivative of the chosen P-gene expression: if the derivative
is positive between measuring times (i.e.~if the concentration of the P-protein
increased), then the cart is pushed right; otherwise, it is pushed left. If
there was no change in its concentration, then the previous
action is repeated.

Another choice lies with the synchronisation between the cart model and the
regulatory model, that is, when to extract the current concentration of the
elected P-protein and feed it to the cart model. As the interval of update for the
cart model is $0.02s$, the interval of measurement of the P-gene
was set to $2000$ time steps. This is however arbitrary, and could become a parameter to
optimise, as it could be set differently for different genomes (some genomes have
slower reactions, others have faster ones).

\subsection{The Evolutionary Algorithm}

The evolutionary algorithm used to evolve the binary genomes was an
evolutionary strategy $(250+250)-ES$: $250$ parents give birth to $250$
offspring, and the best $250$ of all $500$ are used as the new parent
population; a maximum of $50$ iterations were allowed. The only variation
operator used was a simple bit-flip mutation, set to $1\%$ and adapted by the
well-known $1/5$ rule of Evolution
Strategies \cite{rechenberg94a}: when the rate of successful mutations is
higher than $1/5$ (i.e.~when more than 20\% mutation events result in a
reduction of the error measure), the mutation rate is doubled; it is halved in
the opposite case. However, to avoid stagnation of evolution, if the number
of mutation events (i.e.~the number of bits flipped per generation) drops below
$250$, the mutation rate is doubled.

\section{Results and Analysis}
\label{sec:results}

\begin{figure}[ht]
	\begin{center}
		\includegraphics[width=.45\textwidth]{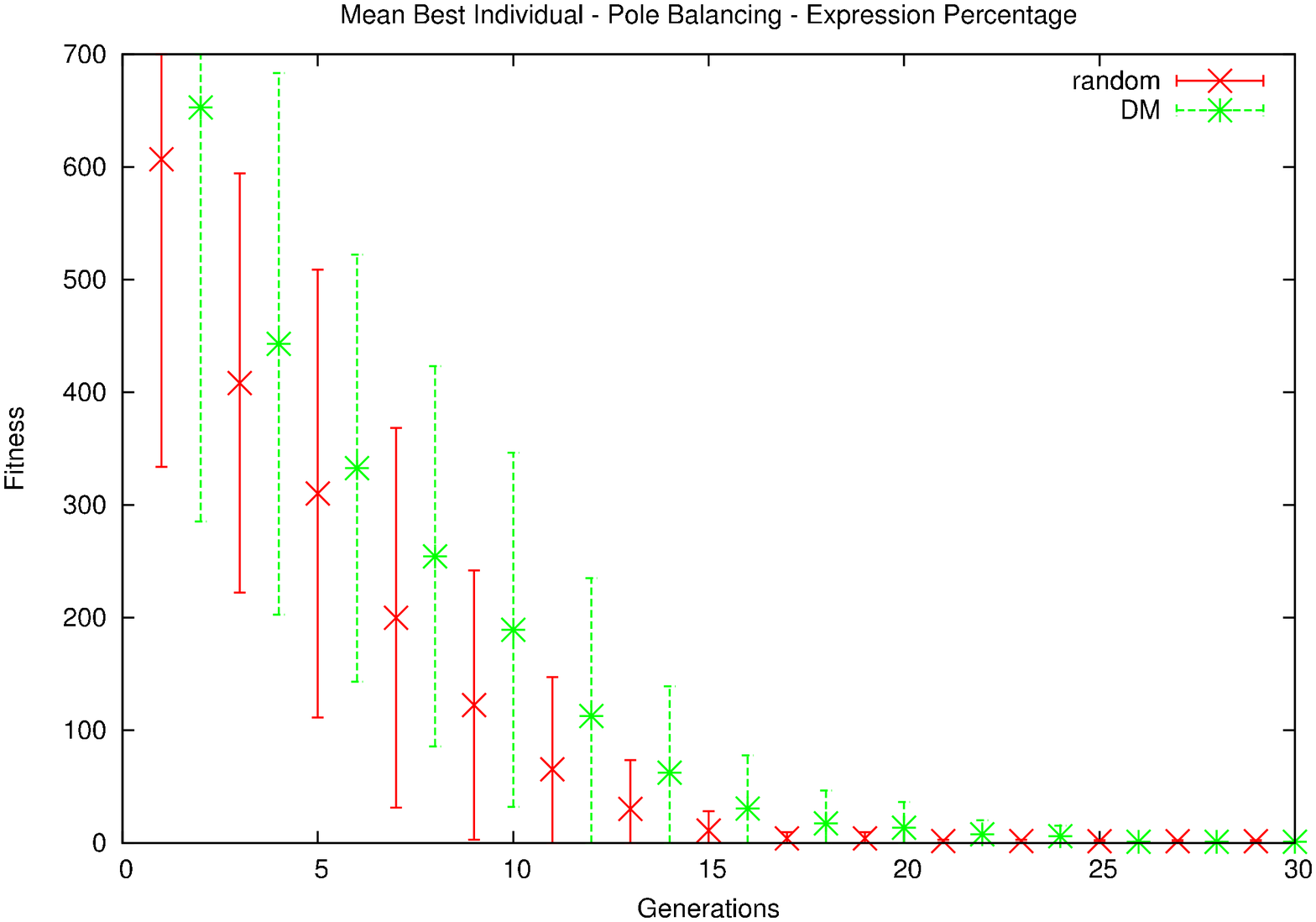}
		\includegraphics[width=.45\textwidth]{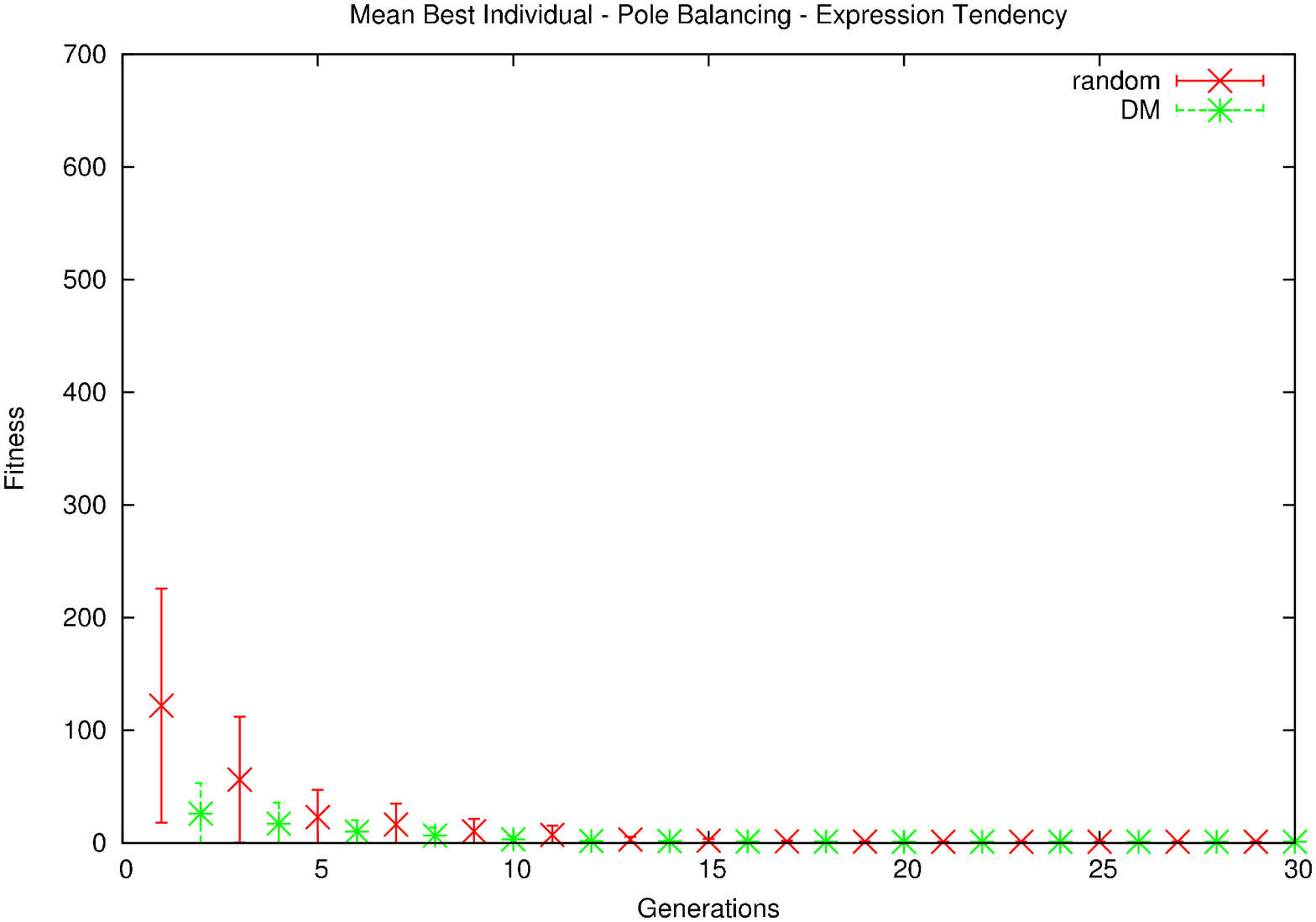}
	\end{center}
	\caption{Mean best individual per generation for the pole-balancing
	problem, when using P-protein concentration (left) or tendency (right).
	All results are averaged across $50$ runs; error bars plot variance
	between runs.}
	\label{fig:poleMBI}
\end{figure}

Fig.~\ref{fig:poleMBI} shows the
average fitness evolution for $50$ independent runs, for both expression
measurement approaches. Both approaches solve the problem quite fast, but it is
obvious that using product tendency gives faster convergence to an optimal
solution. This is an expected result: when using P-protein absolute values, the
concentration of a P-protein has to be fairly close to $50\%$, in order to provide
a solution. However, when using P-protein tendency, the starting concentration of
the P-protein has no influence on the behaviour of the cart.

\subsection{Generalisation Performance}

Whitley et al. \cite{whitley93a} proposed a generalisation test to assert
whether the discovered solution is robust. Once a controller is evolved that can
balance the pole for $120000$ time steps with a random setup, the evolution
cycle is stopped, and this controller is applied to a series of generalisation
tests. These consist of combinations of the four input variables, with their
normalised values set to the following: $0.05$, $0.275$, $0.50$, $0.725$, and
$0.95$. This results in $5^4 = 625$ initial cases. The generalisation score of
the best individual found is thus the number of test cases out of these $625$,
for which the controller manages to balance the pole for $1000$ time steps.

All $50$ runs found solutions for this problem, using either P-protein
concentrations or P-protein tendencies (for both random and DM-genomes).  At the
end of each run, the generalisation test was applied to the best individual in
the population; Table \ref{tab:poleGen} shows the results obtained.

\begin{table*}[ht]
	\begin{center}
		\caption{Generalisation results. Number of successful attempts
		to balance the pole for $1000$ time steps, out of $625$ test
		cases}
		\label{tab:poleGen}
		\begin{tabular}{|c|c|c|c|c|c|c|}
			\hline
			\multicolumn{2}{|c|}{Approach} & Best & Worst & Median & Mean & Std. Dev.\\
			\hline
			Product & random genomes & 422 & 3  & 194 & 202.18 & 110.01\\
			\cline{2-7}
			Percentage & DM-genomes  & 416 & 23 & 237 & 235.68 & 107.85\\
			\hline
			Product & random genomes & 359 &  0 &  63 &  85.82 &  66.99\\
			\cline{2-7}
			Tendency & DM-genomes    & 187 &  7 &  77 &  81.40 &  48.33\\
			\hline
		\end{tabular}
	\end{center}
\end{table*}

The results obtained show little difference between random and DM-genomes.
However, there is a big difference between using P-proteins concentrations or
tendencies, with the former achieving much better results. 
When using product tendency, the concentration of P-proteins can easily
become $0\%$: the previous move is then repeated, and keeps moving the cart 
leftwards. This creates a
disassociation between the product expression and the cart behaviour, which
becomes a handicap when applying the model to some of the harder
generalisation tests.

Note that many of the generalisation tests are unsolvable. After an exhaustive
search of all possible bang-bang solutions up to $60$ steps of simulation, 
$168$ tests were found unsolvable (execution time constrains prevented a deeper
search). This means that an ideal controller can only solve $457$ (or
less) cases. It also shows that the best result found ($422$ tests
solved), although not as high as one of the best in the literature ($446$
solved cases \cite{whitley93a}), is still quite close to the optimum.

\begin{figure}[!ht]
	\begin{center}
		\includegraphics[angle = 270, width=.7\textwidth]{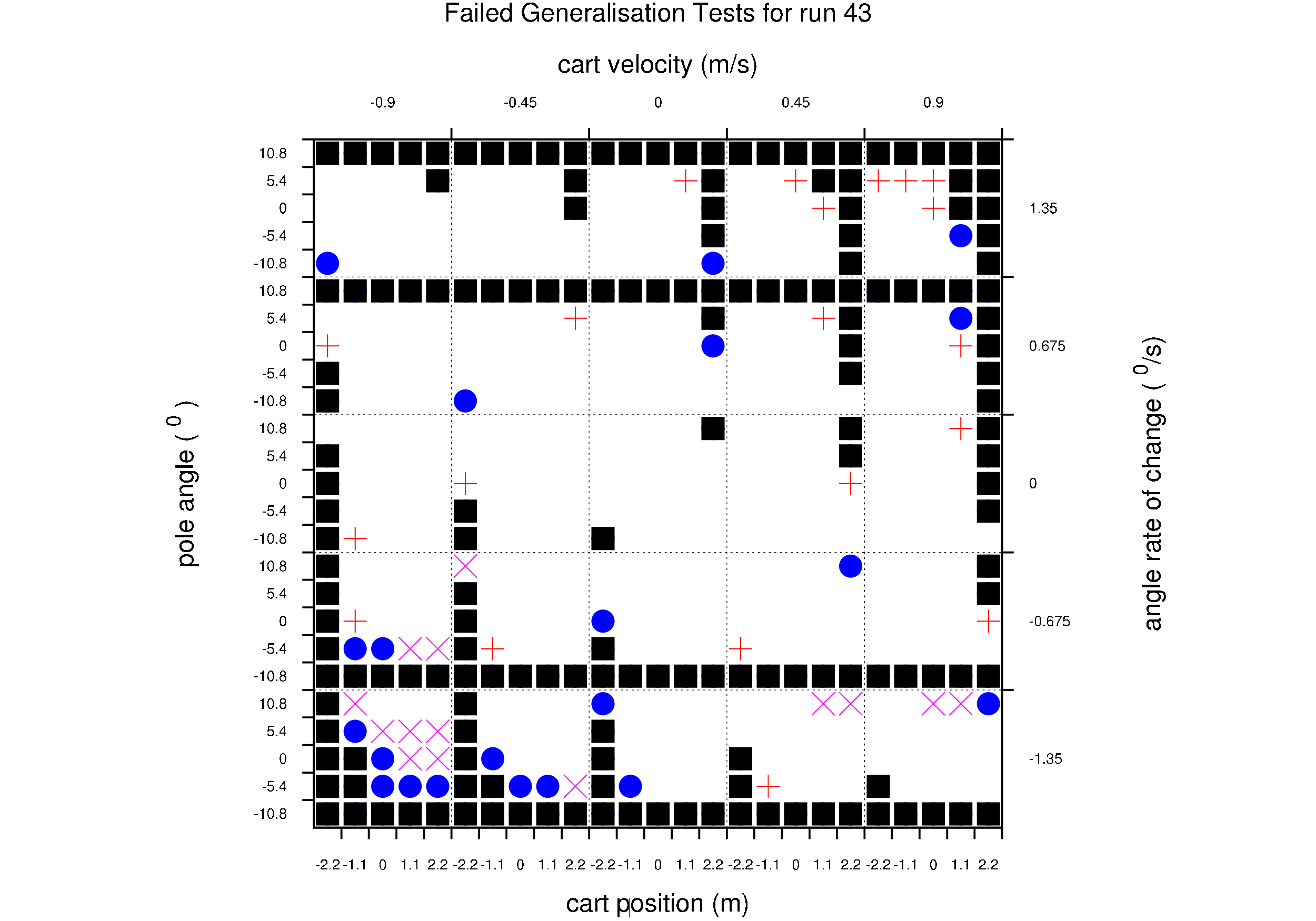}
	\end{center}
	\caption{Generalisation test cases unsolvable at depth $60$ (black
	squares), unsolved by the best random genome ('X'), by the best
	DM-genome ('+'), and by both (filled circles). Both genomes used P-gene
	expression levels.}
	\label{fig:pole60}
\end{figure}

Fig.~\ref{fig:pole60} shows a plot of all the generalisation tests that are not
solvable at depth $60$, and those that are additionally not solved by the best
random and DM-genome. It shows that cases where $\theta$ and $\dot{\theta}$
both take large or small values (i.e.~a large angle in absolute value, together
with a large angular velocity increasing this angle) are unsolvable, and that
both genomes additionally fail on cases that are close to these unsolvable
cases. It is interesting to see however how the unsolved cases of the DM-genome
are mostly symmetric in terms of the matrix of test cases, whereas the random
genome is far more unbalanced. This has to do with the sinusoidal nature of the
controllers generated by random genomes, as can be seen in the next section.


\subsection{Pole balancing behaviour of typical networks}

\begin{figure}[!ht]
	\begin{center}
		$x = 0\ m    \qquad \theta = -5.4\ ^\circ\qquad \dot{x} = 0\ m/s
		\qquad \dot{\theta} = -1.35\ ^\circ/s$\\
		\includegraphics[angle = 270, width=.45\textwidth]{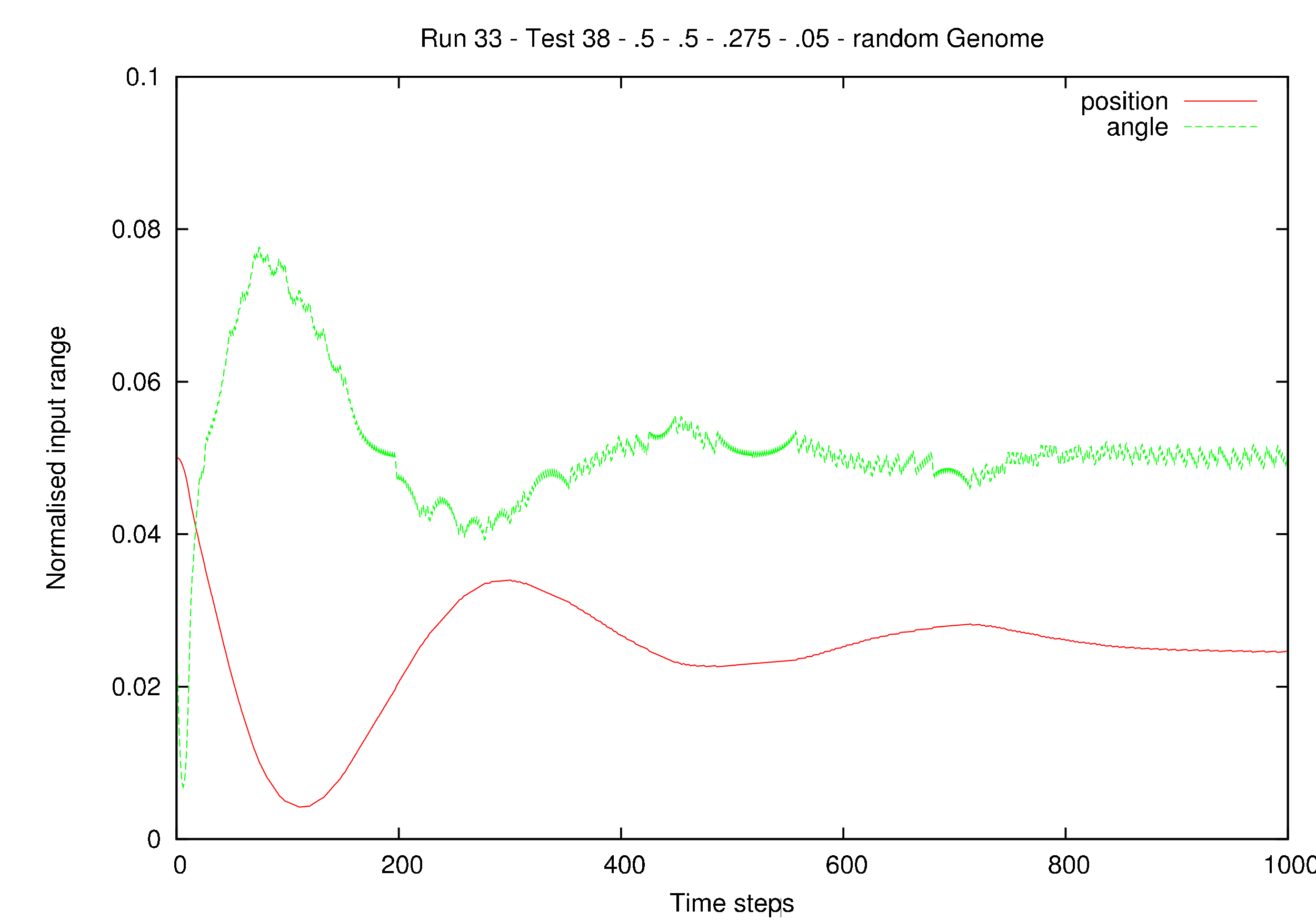}
		\includegraphics[angle = 270, width=.45\textwidth]{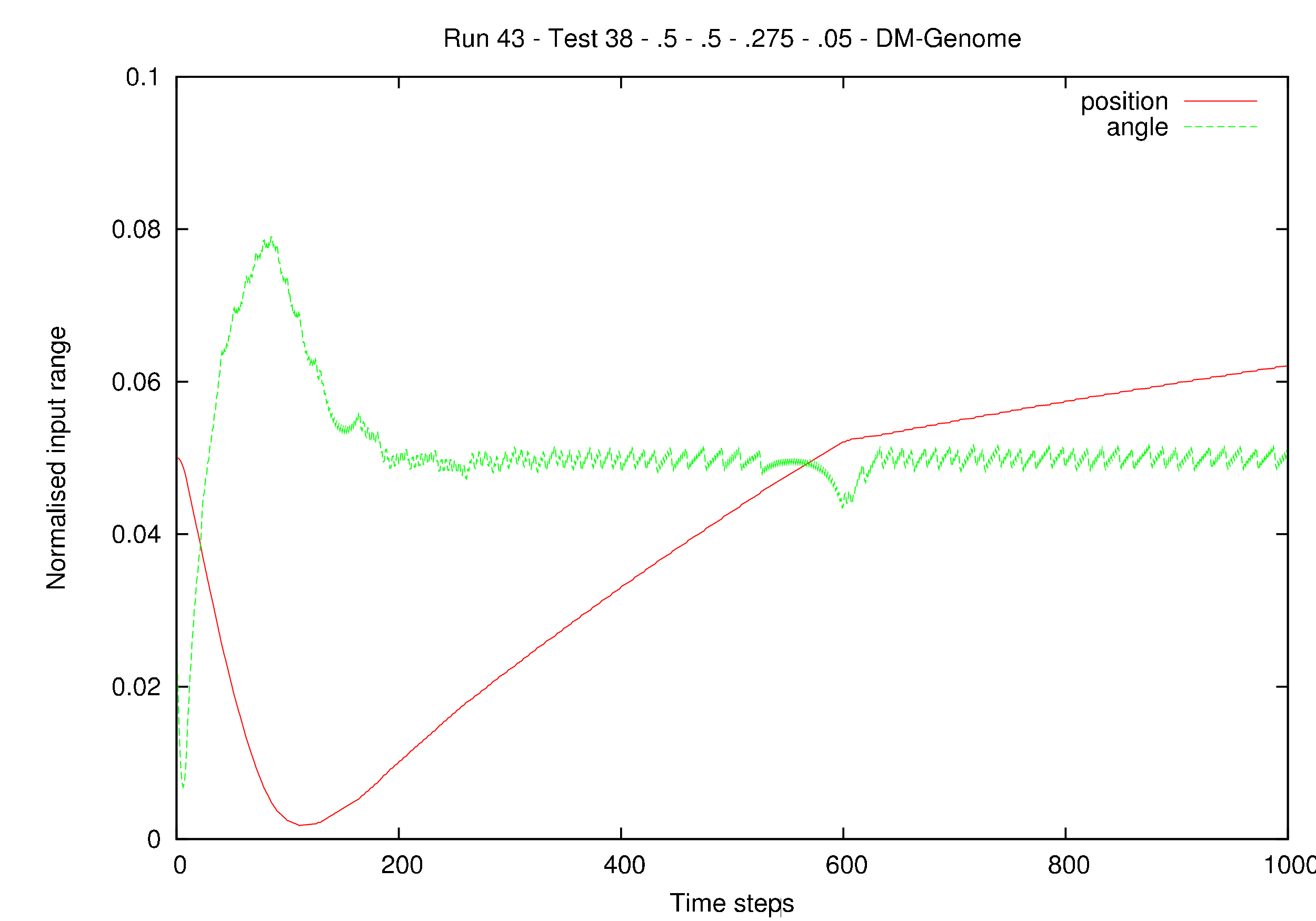}\\\smallskip ~\\
		$x = -1.08\ m\qquad \theta = 5.4\ ^\circ \qquad \dot{x} = 0.45\ m/s
		\qquad \dot{\theta} = 1.35\ ^\circ/s$\\
		\includegraphics[angle = 270, width=.45\textwidth]{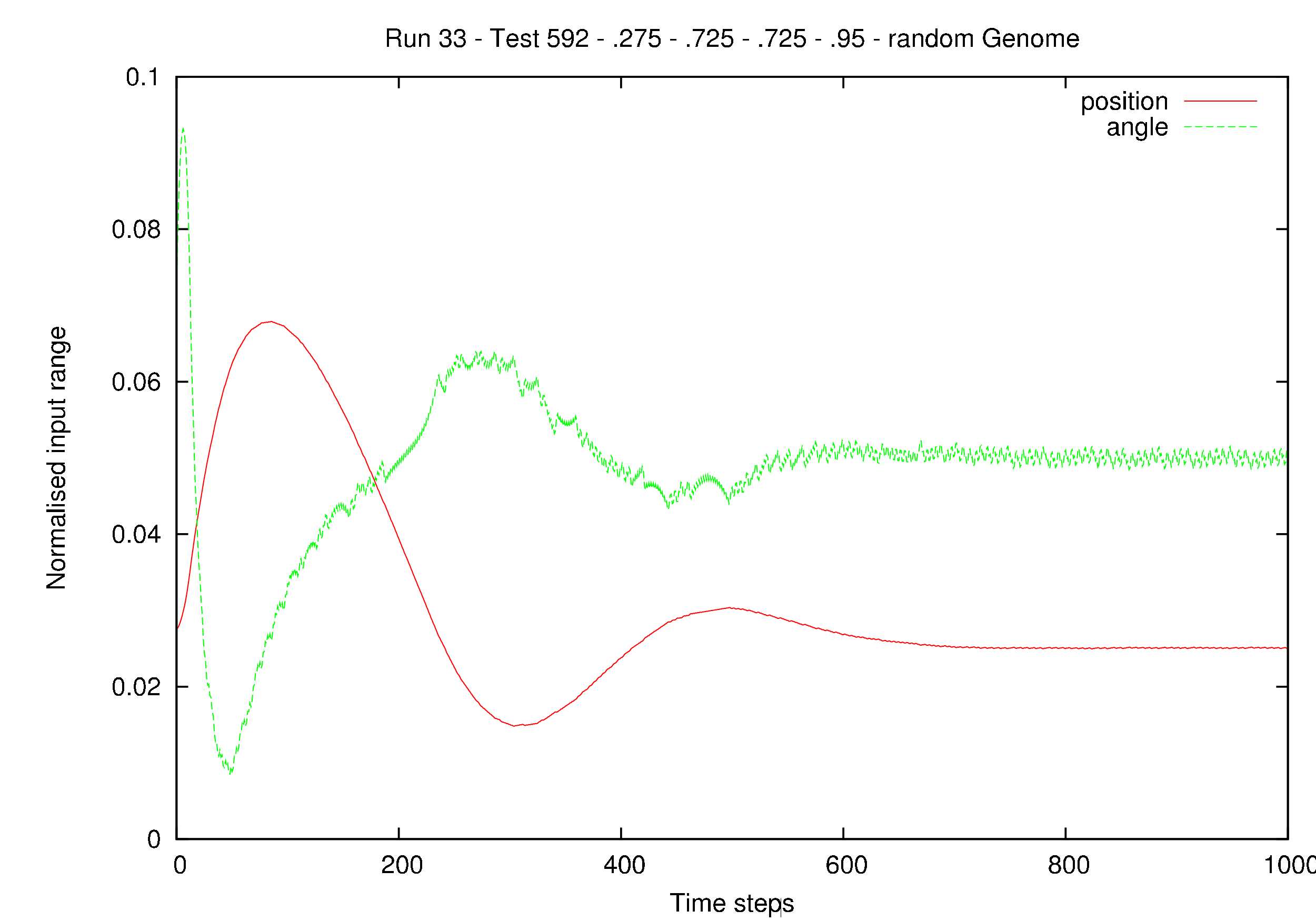}
		\includegraphics[angle = 270, width=.45\textwidth]{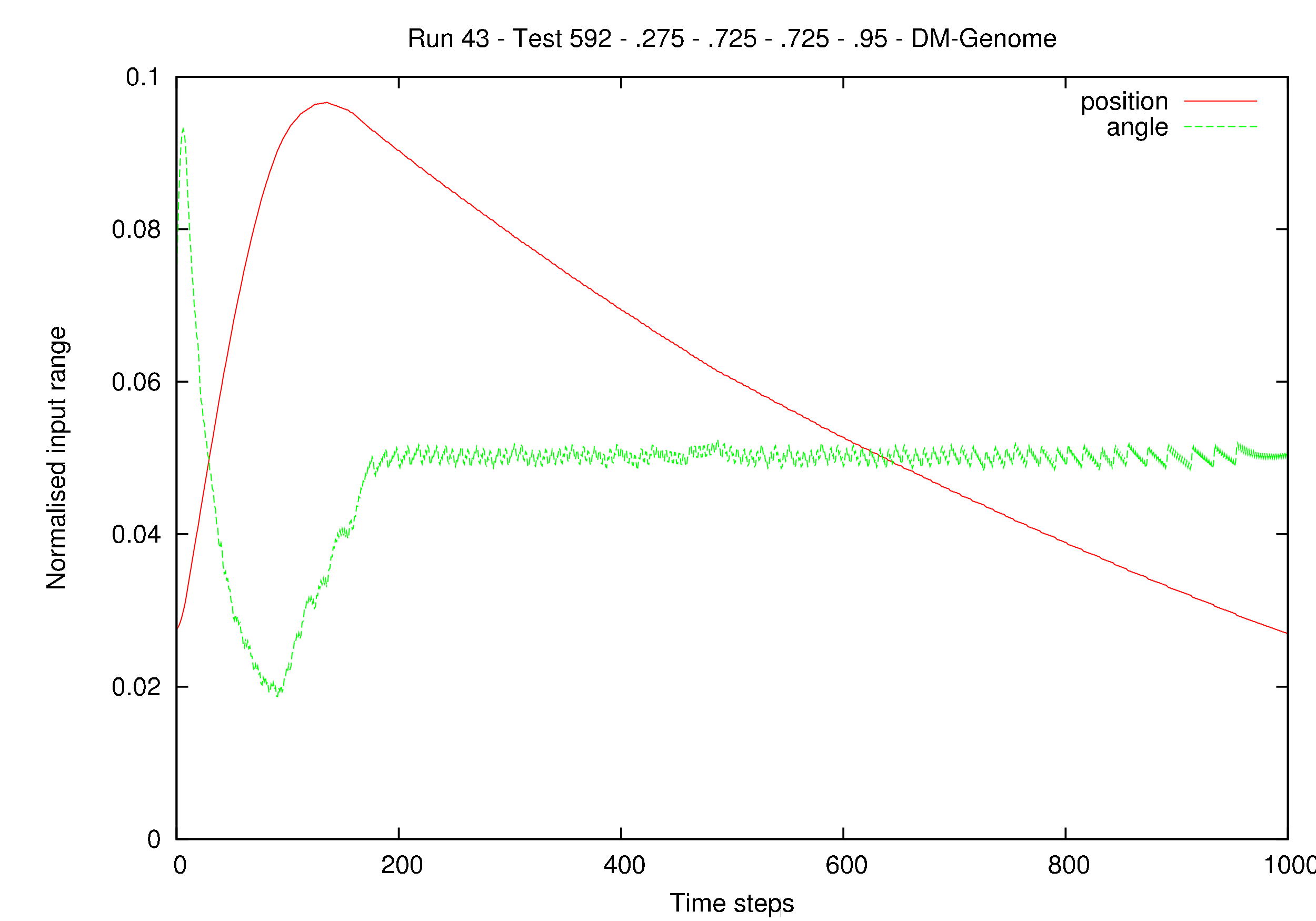}\\
	\end{center}
	\caption{Example progression over time of cart position and pole angle
	for the best random genome (left) and the best DM-genome (right), for
	$2$ generalisation tests.}
	\label{fig:pole-dyn}
\end{figure}

Fig.~\ref{fig:pole-dyn} shows example behaviours of the $2$ best evolved regulatory
models (random and DM, using P-protein concentrations), applied to $3$ different
generalisation cases.

It is interesting to observe the different approaches to solve the same
generalisation test. In particular, one can see how the random genome is quite
sinusoidal in its approach, whereas the DM-genome generates a much more linear
behaviour.

\subsection{Resulting Networks: A typical example}

\begin{figure}[!ht]
	\begin{center}
		\includegraphics[width = .45\textwidth]{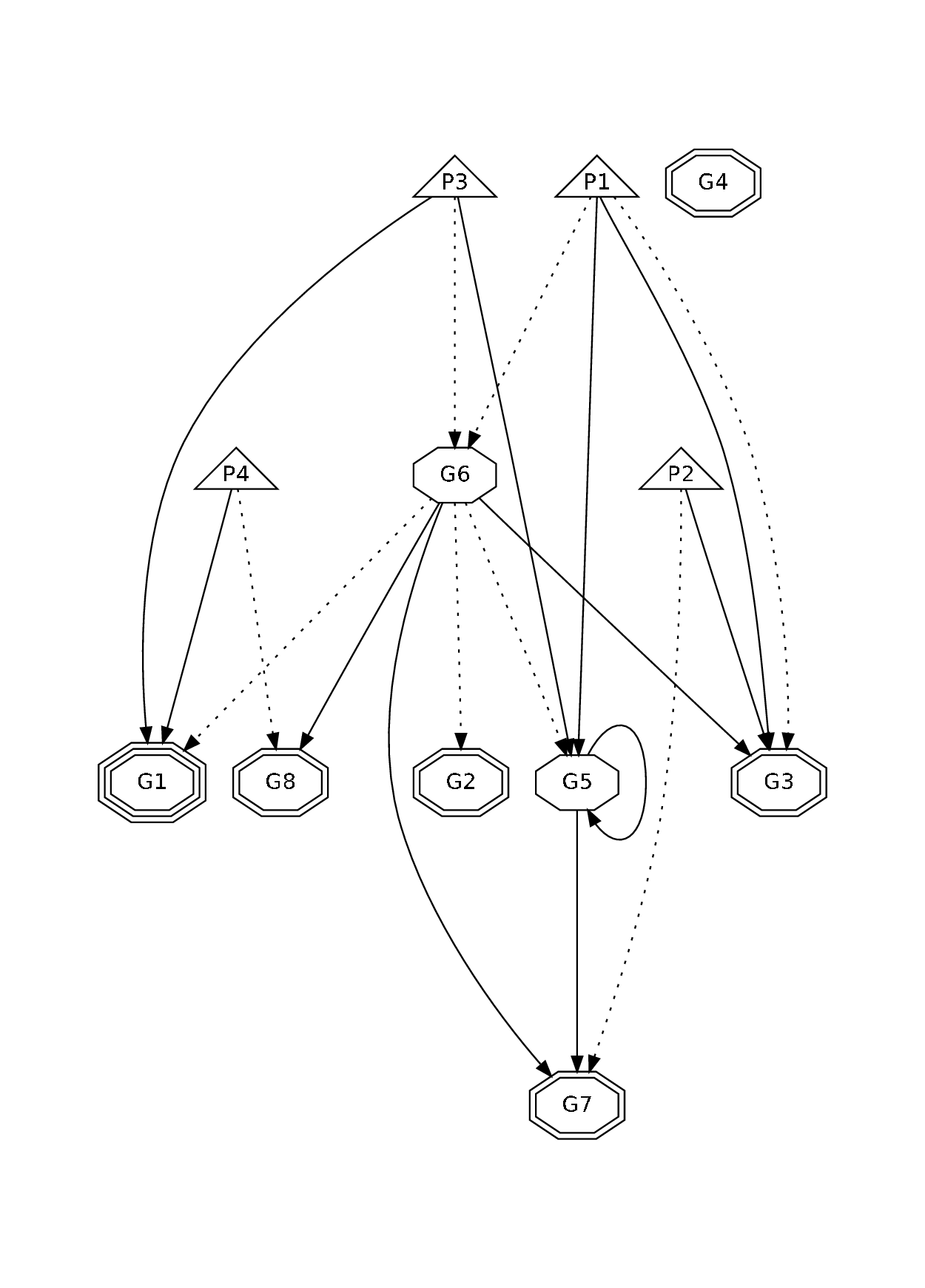}
		\includegraphics[width = .45\textwidth]{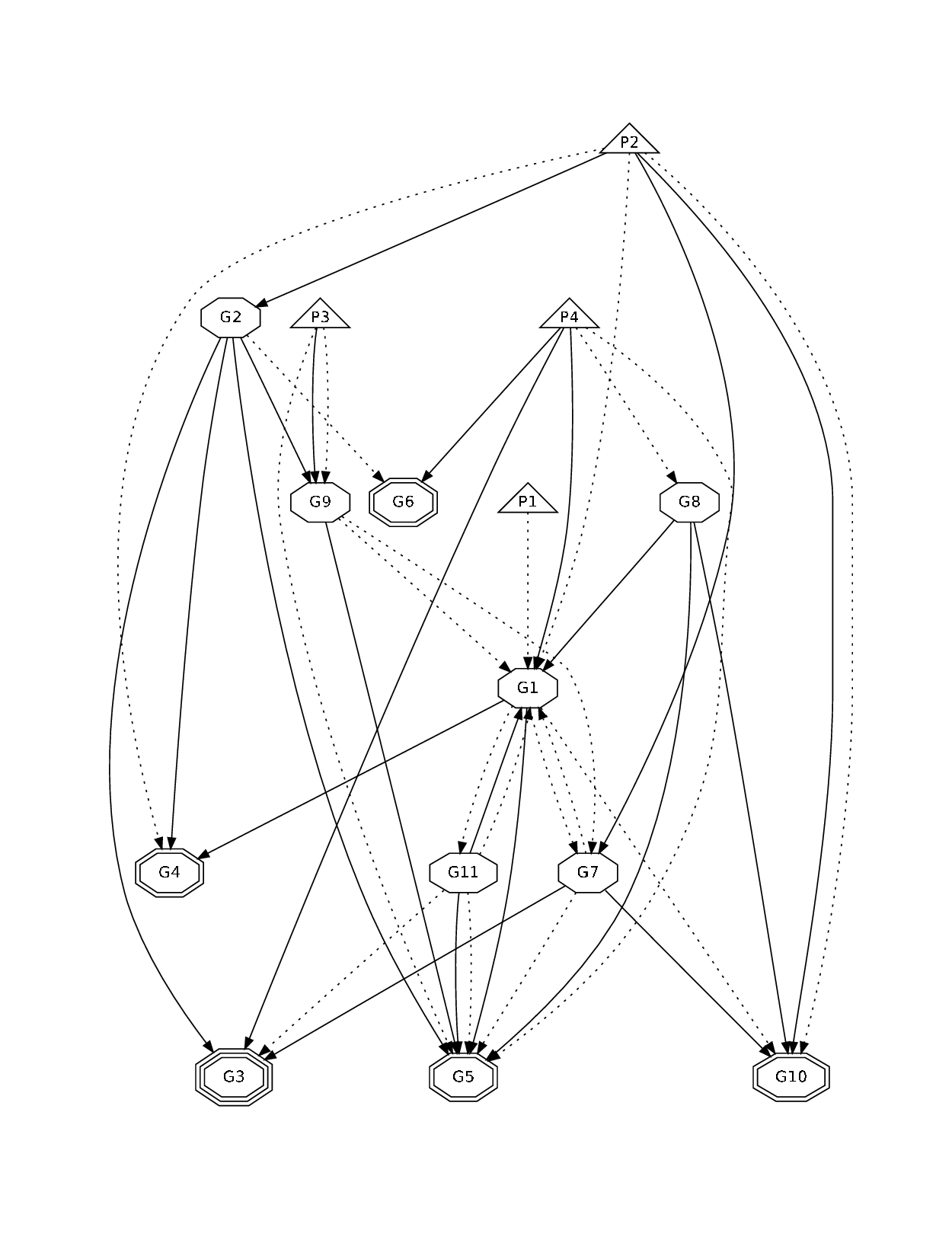}
	\end{center}
	\caption{Regulatory networks extracted from best performing random
	(left) and DM-genome (right). Hexagon nodes represent TF-genes, double
	hexagon nodes represent P-genes, the triple hexagon represents the
	chosen P-gene, and triangles represent the $4$ extra proteins. The
	networks were drawn using a threshold value of $19$.}
	\label{fig:poleex}
\end{figure}

Fig.~\ref{fig:poleex} shows the regulatory networks extracted from the best
performing random and DM-genomes, at a threshold of $19$ (i.e. only connections
with a match larger than 19 are represented, the other ones having a
negligible impact on the regulation -- see Eq.~\ref{eq:e_i}).  Even with such
a low number of genes, one can see that the regulatory interactions are quite
complex. Gene \texttt{G6} seems to act as a central regulatory node on the
random genome, whereas that role is taken up by \texttt{G1} in the DM-genome.
Note also how few connections exist to the chosen P-genes (\texttt{G1} and
\texttt{G3}, respectively); however, the extra protein \texttt{P4}
(representing the rate of change of the pole, $\dot{\theta}$) is directly
connected to these on both genomes. This could very well be a mechanism for
stronger reaction to changes of $\dot{\theta}$, which has been shown to greatly
influence the success rate of a balancing attempt (see Fig.~\ref{fig:pole60}).

\section{Conclusions}
\label{sec:conclusions}

One of the main objectives of this paper was to investigate the possibility of
using GRNs as a new representation for program synthesis through Genetic Programming.
Our motivation was that today's mainstream Evolutionary Computation (EC) approaches
are by and large crude simplifications of the biological paradigm of natural evolution, not taking into
account many advances of biological knowledge in the recent past
\cite{banzhaf06a}. The artificial GRN model used \cite{banzhaf03a} presents an
interesting balance between biological accuracy and computational potential,
and was proposed as a good basis to introduce more accurate biological basis
for EC.

The results obtained show that there is a clear computational potential within
the model; it should therefore be possible to use other similar models as basis
for EC techniques.

The adaptation of such models to EC is not straightforward. As these are mostly
complex systems, a thorough comprehension of their exact dynamics is often not
possible. The choice of how to encode inputs and outputs is also not a simple
issue, and can greatly influence their computational potential.

Another key issue is the execution speed. While their biological equivalent
systems are extremely fast, at the moment these computer models are somewhat
slow, and the model used here is no exception. In order to accelerate the
regulatory reactions, several tricks were used, such as adapting the sampling
time of the differential equation (the $\delta$ parameter), and parallelization
by distributing the evaluation of genomes across a cluster -- the resulting
average execution time of a single run was around $25$ minutes, when executing
the code on $8$ recent machines running in parallel. Of course, a fascinating
possibility to overcome this issue would be to synthesize the resulting GRN
into biological medium.

Regarding this problem, some parameters could be optimized (e.g.~by evolution).
First, the signature and concentration of the extra proteins: a deeper
understanding of their influence on the regulatory process is necessary; it
could very well be that their influence is far too strong for the moment.

Second, the synchronisation between the biological and physical models. As
mentioned before, different models have different reaction times (for example,
stabilization times for genomes of this size may go from a few thousand
iterations up to hundreds of thousands); each genome would therefore need to
tune its synchronisation period individually.

Future work will now focus on extensive testing of the new extended GRN model
on various problem domains. The most promising ones seem to be dynamic control
problems, as these might profit the most from the remarkable dynamic properties
of the model. But the flexibility of this representation allows one to imagine
more GP-like approaches. For example, even though only 2 types of proteins were
used, a lot more could be introduced - and potentially represent the equivalent
of GP functions or terminals. The change of their concentrations over time
could then represent priorities of execution, or even probabilities. Work is
under way to explore these new avenues of investigation.


\end{document}